\documentclass{article}

\usepackage{arxiv}

\usepackage[utf8]{inputenc} 
\usepackage[T1]{fontenc}    
\usepackage{hyperref}       
\usepackage{url}            
\usepackage{booktabs}       
\usepackage{amsfonts}       
\usepackage{nicefrac}       
\usepackage{microtype}      
\usepackage{lipsum}		
\usepackage{graphicx}
\usepackage{natbib}
\usepackage{doi}
 \usepackage{amsmath}
 \usepackage{multirow} 

\title{Rethinking Nonlinearity: Trainable Gaussian Mixture Modules for Modern Neural Architectures}

\author{ Weiguo Lu \\
	Guangdong Institute of Intelligence\\Science and Technology\\
	\texttt{luweiguo@gdiist.cn} \\
	\And
	Gangnan Yuan\thanks{Co-corresponding Authors} \\
	Great Bay University\\
    ~~\\
	\texttt{gnyuan@gbu.edu.cn} \\
	\AND
	Hong-kun Zhang \\
    Great Bay University\\
	University of Massachusetts at Amherst \\
	\texttt{hongkunz@umass.edu} \\
	\And
	Shangyang Li*\\
	Guangdong Institute of Intelligence\\Science and Technology \\
	\texttt{lishangyang@gdiist.cn} \\
}

\renewcommand{\shorttitle}{\textit{arXiv} Template}

\hypersetup{
pdftitle={A template for the arxiv style},
pdfsubject={q-bio.NC, q-bio.QM},
pdfauthor={David S.~Hippocampus, Elias D.~Striatum},
pdfkeywords={First keyword, Second keyword, More},
}

\begin{document}
\maketitle

\begin{abstract}
Neural networks in general, from MLPs and CNNs to attention-based Transformers, are constructed from layers of linear combinations followed by nonlinear operations such as ReLU, Sigmoid, or Softmax. Despite their strength, these conventional designs are often limited in introducing non-linearity by the choice of activation functions. In this work, we introduce Gaussian Mixture-Inspired Nonlinear Modules (GMNM), a new class of differentiable modules that draw on the universal density approximation Gaussian mixture models (GMMs) and distance properties (metric space)  of Gaussian kernal. By relaxing probabilistic constraints and adopting a flexible parameterization of Gaussian projections, GMNM can be seamlessly integrated into diverse neural architectures and trained end-to-end with gradient-based methods. Our experiments demonstrate that incorporating GMNM into architectures such as MLPs, CNNs, attention mechanisms, and LSTMs consistently improves performance over standard baselines. These results highlight GMNM’s potential as a powerful and flexible module for enhancing efficiency and accuracy across a wide range of machine learning applications.
\end{abstract}


\section{Introduction}
Many classical machine learning methods and contemporary neural network models—such as variational inference, Gaussian processes, kernel methods, expectation–maximization algorithms, t-distributed stochastic neighbor embedding, autoencoders, and generative diffusion models—are inherently connected to the exponential family \cite{murphy2012machine,dempster1977maximum,attias1999variational,blei2017variational,hinton2002stochastic,kingma2013auto,song2020score,sohl2015deep,ho2020denoising,song2019generative}. Although Gaussian Mixture Models (GMMs) are highly effective for modeling multi-modal distributions, neural networks utilizing GMMs remain underexplored. Early concepts of Gaussian mixture networks date back to 1989 \cite{alba1999growing}, with recent works introducing advanced techniques such as Gaussian mixture convolutions and Gaussian mixture conditioning \cite{alba1999growing,tsuji1999log,celarek2022gaussian,lu2025diffusion}. Contemporary neural architectures typically rely on linear transformations followed by nonlinear activations, leaving the potential advantages of exponential-family-based mixtures largely untapped. To address this gap, we propose an intuitive yet powerful integration of Gaussian mixtures into neural network architectures, demonstrating significant performance improvements in our experiments.

This work is motivated by foundational theoretical insights, similar to how Multi-Layer Perceptrons (MLPs) emerged from the universal approximation theorem\cite{hornik1989multilayer}, subsequently inspiring convolutional layers, Long Short-Term Memory (LSTM) units, attention mechanisms, and recently, Kolmogorov–Arnold Networks (KANs) \cite{liu2024kan}. Specifically, our approach is inspired by \cite{villani2008optimal} indicating that mixture models approximate arbitrary probability densities under Wasserstein distance, and the universal approximation capability of radial basis function (RBF) networks shown by \cite{park1991universal}. Building upon these results, we introduce the Gaussian Mixture-Inspired Nonlinear Modules (GMNM), a neural architecture derived from a reformulation of the GMM.

\textbf{The key idea of GMNM is to reinterpret GMMs as neural components, allowing their beneficial properties to enhance modern deep learning frameworks.} By relaxing traditional probabilistic constraints, we transform GMMs into flexible universal function approximators beyond density estimation tasks. Additionally, the intrinsic nonlinearity introduced by the Mahalanobis distance within Gaussian functions provides distinct representational advantages compared to conventional neural network activations.

Specifically, GMNM removes constraints on mixture weights, converting GMMs into versatile function approximators optimized via gradient descent instead of specialized algorithms such as Expectation-Maximization (EM) or variational Bayesian methods. Moreover, GMNM introduces a simple yet effective approach to handle covariance matrices in high dimensions, addressing the usual difficulties associated with learning positive definite covariance matrices. Unlike traditional RBF networks, GMNM is not restricted to isotropic covariance, enabling robust modeling of complex data correlations and distributions.

We extensively evaluate GMNM across various tasks, including two-dimensional function fitting and partial differential equation (PDE) approximation. Our results consistently demonstrate GMNM's superior performance relative to standard MLPs and KANs. Furthermore, integrating GMNM into established architectures—such as Convolutional Neural Networks (CNNs), with and without attention mechanisms, and LSTMs—leads to notable performance improvements, underscoring GMNM's applicability and effectiveness.

With the presentation of GMNM, we thus provide the following scientific contributions in this work:
\begin{itemize}
\item We demonstrate experimentally that relaxing probabilistic constraints transforms Gaussian Mixture Models into powerful universal function approximators, surpassing conventional architectures like MLPs and KANs.
\item We propose the GMNM architecture, enabling gradient-based training of Gaussian mixtures within neural networks and efficiently modeling high-dimensional data correlations.
\item We show that integrating GMNM as a modular component significantly enhances existing neural architectures, including attention-equipped CNNs and LSTMs.
\end{itemize}

\section{Related Works}
\textbf{Radial Basis Function Networks}, introduced by \citeauthor{lowe1988multivariable}, approximate functions through linear combinations of radially symmetric activation functions that respond to the distance between inputs and learned centers. While theoretically capable of universal approximation \cite{park1991universal}, RBF networks face significant limitations, such as limited scalability for high-dimensional problems. Similar to the structure of RBF networks, our work can prove its universal approximation by the same theory, but we achieve a better adaptation to high-dimensional problems by a different construction of the mean and covariance.

\textbf{Attention Mechanism,} first popularized in neural machine translation by \cite{chorowski2015attention}, enable models to selectively focus on relevant parts of input data when generating outputs. This capability has proven transformative across multiple domains, serving as the foundation for Transformer architectures (\cite{vaswani2017attention}) that have revolutionized natural language processing and, increasingly, computer vision and multimodal learning. In this work, similar to the attention  mechanism, the constructed GMNM module is embedded into the CNN to realize more types of tasks.

\textbf{Kolmogorov-Arnold Networks,}  introduced by \cite{liu2024kan}, implement the Kolmogorov-Arnold representation theorem through a neural architecture that approximates multivariate functions as compositions of univariate functions. The performance of KAN and MLP has its own pros and cons in different tasks (\cite{yu2024kan}). Some new perspectives \cite{li2024kolmogorov} suggest that KAN based on Gaussian kernel is intrinsically the same as RBF networks. 

\textbf{Gaussian Mixture Models} represent complex probability distributions as weighted sums of Gaussian components, enabling flexible density estimation across diverse data types \cite{reynolds2009gaussian}. GMM is commonly used for clustering, density estimation, and anomaly detection \cite{wan2019novel,lu2025efficient,wang2023hyperspectral}. Each Gaussian distribution is defined by its mean and covariance. The advantage of GMM is its ability to model complex, multi-peaked data distributions. While powerful for clustering and density estimation, traditional GMM face challenges with high-dimensional data, where parameter estimation becomes ill-conditioned and the curse of dimensionality limits their effectiveness without structural constraints or dimensionality reduction techniques.

\textbf{Gaussian Mixture Networks.} Current deep learning-based Gaussian mixture models typically learn the parameters of GMM using deep neural networks, but most approaches assume the data strictly follows a Gaussian mixture distribution and rely on the Expectation-Maximization algorithm for training \cite{celarek2022gaussian, el2023deep, viroli2019deep}. These constraints lead to one biggest limitations: the EM algorithm is not close to any form of a loss function, and it is not a gradient-based method, which is not compatible with modern neural networks. This work is released from these limitations and can be deployed on a wider range of tasks and is compatible with other neural network structures.

\section{Gaussian Mixture-Inspired Nonlinear Modules}\label{sec:GMNM}

The conventional Gaussian mixture model is a type of mixture distribution that assumes all data points are generated from a mixture of a finite number of Gaussian distributions. The probability density function $g(x)$ of a GMM  is a linear combination of the Gaussian distributions. For each weight $\pi_n \geq 0$ and $\sum_{n=1}^{N}\pi_n =1$,
\begin{eqnarray}\label{GMMdensity}
    g(x)=\sum_{n=1}^{N}\pi_n \phi(x;M_n, \Sigma_n),
\end{eqnarray}
where $\phi$ is density function of Gaussian distributions, $x\in \mathbb{R}^D$, mean vector $M_n\in \mathbb{R}^D$, covariance matrix $\Sigma_n\in{\mathbb{R}^{+}}^D\times {\mathbb{R}^{+}}^D$ and $N,D\in \mathbb{N}^+$. 

One of the key motivations for build Gaussian Mixture Models (GMMs) into a network-wise model lies in their universal approximation capabilities. Prior work (\citet{villani2008optimal,lu2025efficient}) has shown that GMMs can approximate arbitrary probability density functions with arbitrary precision under the Wasserstein distance. If the constraint requiring the function to be a valid probability density is relaxed, i.e., let $\pi \in \mathcal{R}$, then the GMM is equivalent to an RBF based on a Gaussian kernel. \cite{park1991universal} proves that Eq. (\ref{GMMdensity}) is dense in $L^1 (\mathbb{R}^d)$ under the above condition, that is, the GMM without coefficient restrictions satisfies the universal approximation theorem.

The architecture of a GMM naturally exhibits a network-like structure through its computation. However, in high-dimensional settings, managing the covariance matrix $\Sigma$ becomes a critical challenge. When an isotropic covariance is used—assuming independence across dimensions—the model is unable to learn correlations between dimensions. To enable the learning of such correlations, the covariance matrix must remain positive definite, necessitating specialized treatment within the system.
 
In this work, we introduce the Gaussian Mixture-Inspired Nonlinear Modules (GMNM), which provides an alternative strategy to approximate the quadratic term \((x-\mu)^T\Sigma^{-1}(x-\mu)\) in the multivariate Gaussian distribution. Consider a multivariate Gaussian distribution \(G(x)\):

\begin{equation}\label{eq-gaussian}
  G(x) = \frac{1}{(2\pi)^{d/2}\sqrt{\det(\Sigma)}} \exp \bigl( -0.5 \,(x - \mu)^T \Sigma^{-1} (x - \mu) \bigr),
\end{equation}

where \(\Sigma\) is a positive definite covariance matrix. Consequently, its inverse \(\Sigma^{-1}\) must also be positive definite, ensuring \((x - \mu)^T \Sigma^{-1} (x - \mu) \geq 0\). In a back-propagation setting that directly parameterizes \(\mu\) and \(\Sigma^{-1}\), it is challenging to maintain \(\Sigma^{-1}\) as positive definite at every training step.

\begin{figure}[h]
  \centering
  \includegraphics[scale=0.24]{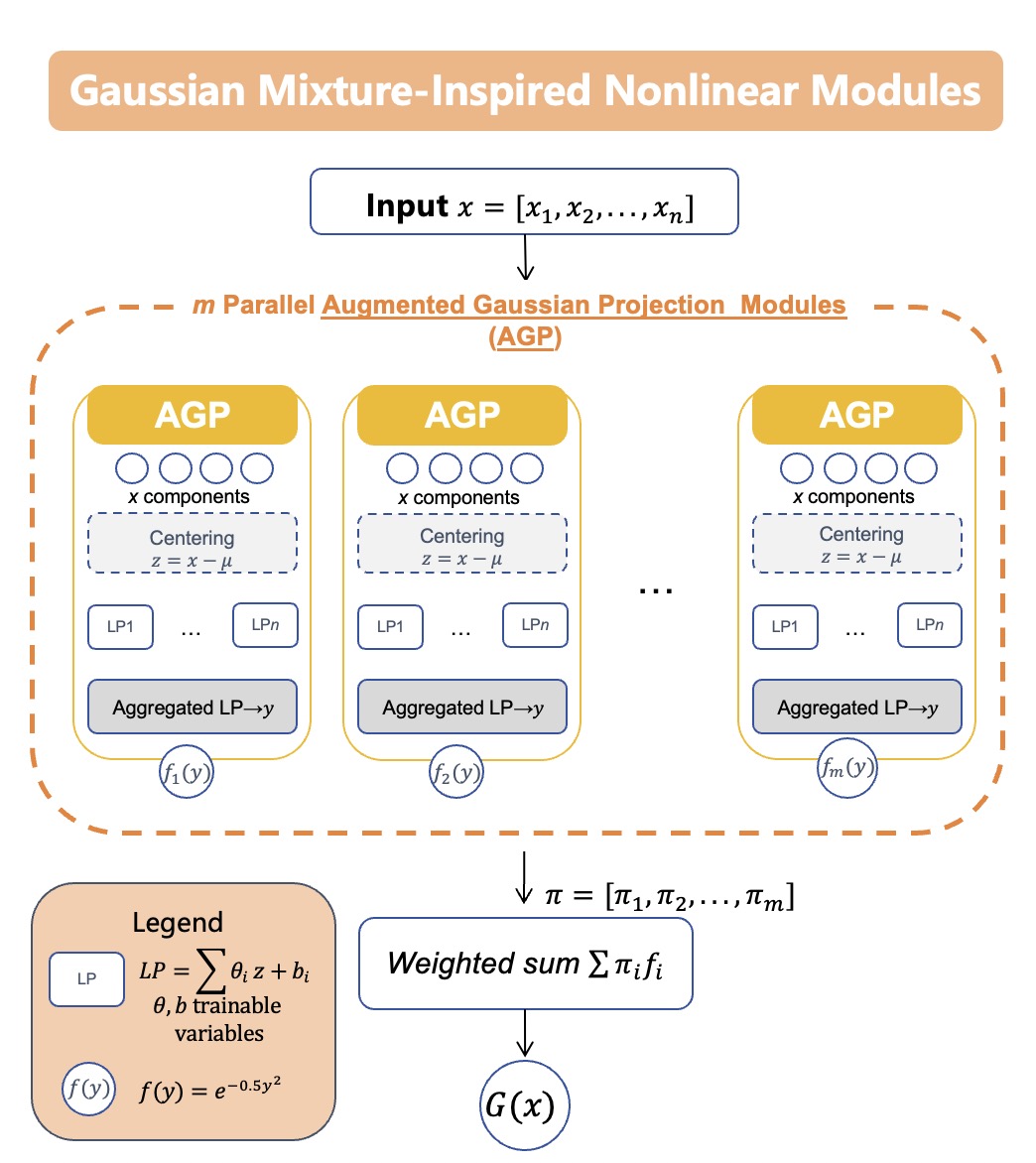}
  \caption{Architecture of the proposed GMNM.  The input \(\mathbf{x} = [x_1, x_2, \dots, x_d]\) is processed by \(m\) parallel Augmented Gaussian Projection (AGP) modules. 
  Each AGP applies centering \(\mathbf{z} = \mathbf{x} - \boldsymbol{\mu}\) and then uses multiple linear projections (LP) to form an aggregated scalar output \(y\), 
  which is mapped through \(f(y) = e^{-0.5 \, y^2}\). 
  The outputs \(f_i(y)\) from individual AGP modules are linearly combined with weights \(\{\pi_i\}\) to produce the final mixture output \(G(\mathbf{x})\).}
  \label{agmg-arch}
\end{figure}

To circumvent the direct inversion of \(\Sigma\), we propose replacing \((x-\mu)^T \Sigma^{-1} (x-\mu)\) with a flexible pair of linear projections and a squaring function, as illustrated in Figure~\ref{agmg-arch}. Specifically:

\begin{equation}
\begin{gathered}
    z = x - \mu,\\
    \text{LP}_{n}(z) = \sum a_{n} \, z_n + b_{n},\\
    y = \sum\nolimits_{n} \alpha_{n}\,\text{LP}_n(z) + \beta, \\
    f(y) = \exp(-0.5 \, y^2),
\end{gathered}\label{eq3}
\end{equation}

where \(x\) and \(\mu\) are \(1 \times d\) vectors, and \(n\) typically equals \(d\). Each \(\text{LP}_n\) is a linear projection, and after two successive projections (with intermediate squaring or exponentiation), the resulting scalar \(y\) is mapped through \(f(y)\) to emulate the Gaussian exponential term.

The motivation behind Eq.~\eqref{eq3} is to capture a structure that approximates the Mahalanobis distance \((x-\mu)^T\Sigma^{-1}(x-\mu)\) while remaining flexible enough to ensure positivity without having to strictly enforce positive definiteness at each training step. In the supplementary materials, we prove that this structure is generalized equivalent to the Mahalanobis Distance.

Each \textbf{Augmented Gaussian Projection (AGP)} produces a single output \(f_i(y)\). By combining these outputs through a set of learnable coefficients \(\pi_i\), we obtain the overall mixture output \(G(x)\). Although \(G(x)\) resembles a Gaussian mixture density, we do not impose the normalization constraint \(\sum \pi_i = 1\). Consequently, we also omit the constant factor \(\frac{1}{(2\pi)^{d/2}\sqrt{\det(\Sigma)}}\) for simplicity, which benefits back-propagation by avoiding gradients with respect to \(\det(\Sigma)\). Specifically:

\begin{itemize}
  \item \textbf{Relaxed normalization:} For general function fitting, the integral \(\int_{-\infty}^{\infty} G(x)\,dx = 1\) is not strictly required, so the determinant term and the factor \((2\pi)^{-d/2}\) need not be preserved.
  \item \textbf{Simplified training:} Because the parameters \(\pi_i\) are not constrained to sum to 1, the normalizing constant can be absorbed into these weights without affecting the overall model expressiveness.
\end{itemize}

\section{Experiments: Approximation Property}\label{experiments}

We design GMNM as a Neural Network module but fundamentally speaking, this module on itself is also a model as well as a function approximator. First and foremost, we assess the function approximation properties of GMNM in this section and exam other properties with neural network in next section. As both MLP, KAN and GMMs are supported by universal approximation theorems, it is crucial to quantitatively compare their performance in representing complex functions. The experimental settings are as follows: (1) two-dimensional function approximation, (2) solving partial differential equations. Through these experiments, we aim to provide a comprehensive analysis of GMNM's approximation accuracy. All experiments were conducted on Google Colab using a T4 GPU.

\subsection{GMNM for Function Fitting}\label{sec:GMNM-funcfit}

We first study GMNM in an unconstrained setting where the Gaussian components are used purely as non-linear basis functions rather than probability densities.  
Besides the 1-D results in supplementary materials, we report here a 2-D benchmark that lets us gradually raise the level of difficulty.

\begin{figure}[h]
  \centering
  \includegraphics[scale=0.21]{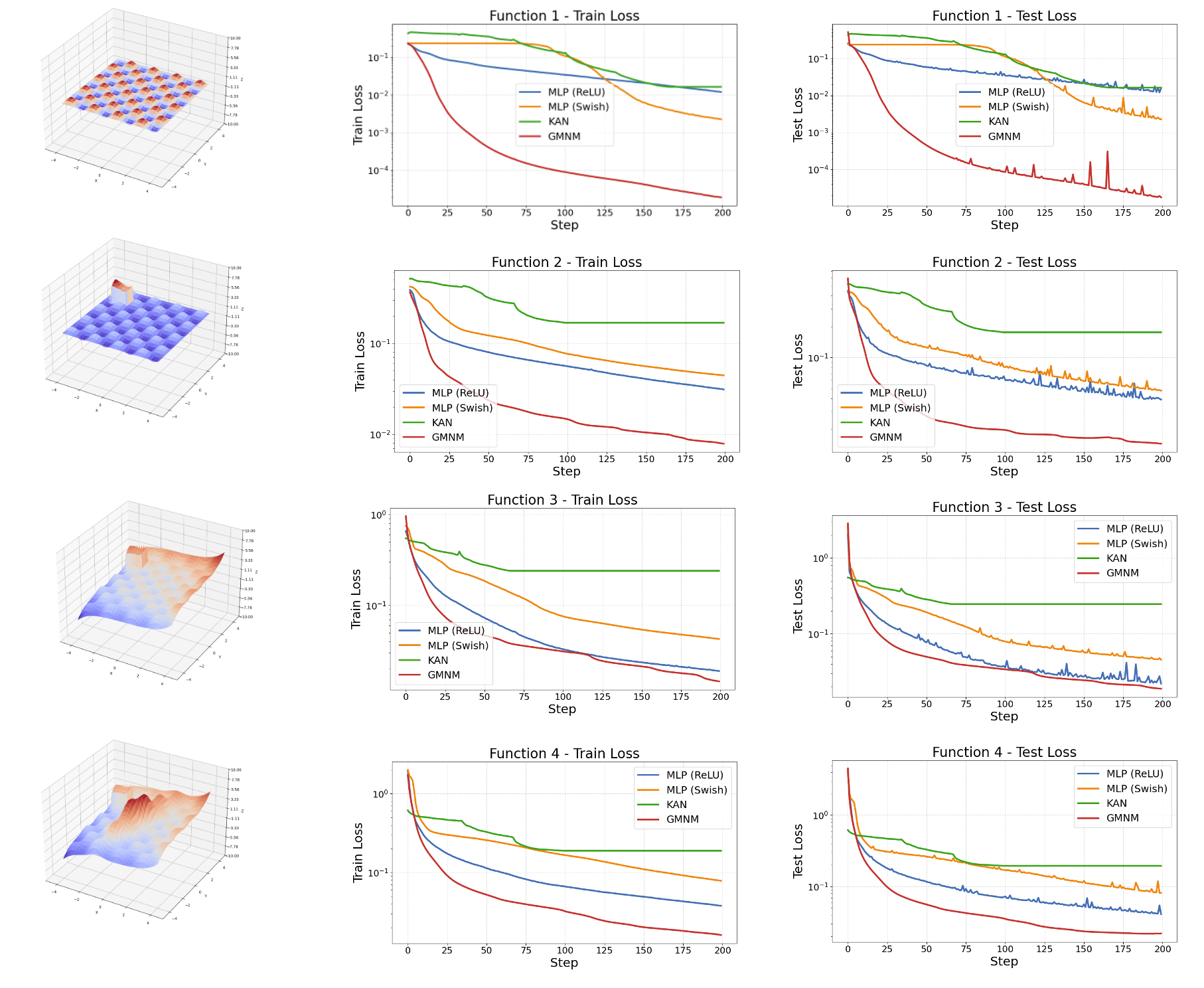}
  \caption{2D function fitting results for progressively more complex tarrfaces (left),
  training losses (middle), and test losses (right). 
  Again, GMNM (red) outperforms KAN (green) and MLPs (blue and orange).}
  \label{fig3}
\end{figure}

Set of 2D target functions:
\begin{align*}
T(\mathbf{x}) &= \sin(\pi x_1)\sin(\pi x_2)
    - a\,\bigl(\sinh(x_1)+\sinh(x_2)\bigr) \\
    &\quad + b\,U(x_1,x_2) + c\,\mathcal{N}_{\mu,\Sigma}(x_1,x_2), \\[1ex]
U(x_1,x_2) &=
    \begin{cases}
        \dfrac{1}{1.5} & x_1\in[-3.5,-2.0],\, x_2\in[2.0,3.0],\\
        0 & \text{otherwise},
    \end{cases} \\[1ex]
\mu &= [0,0], \quad 
\Sigma = \begin{pmatrix} 0.6 & 2.0 \\ 1.1 & 3.9 \end{pmatrix}.
\end{align*}
where again \((a,b,c)\) control the complexity of the function, they are $[0,0,0],[5.0,0,0],[5.0,0.1,0],[5.0,0.1,40.0]$. Figure~\ref{fig3} shows examples of the 2D target surfaces (left line), along with corresponding training-loss (middle line) and test-loss (right line) curves.
The baseline methods are KAN and MLP with two different activation functions, implement details and larger figures can be found in the supplementary materials. Since the RBF network performed weaker than the MLPs, we did not add it to baseline.
Key observations:

• Fast convergence.  GMNM descends almost monotonically, whereas the other methods plateau early.  

• Robust generalisation.  KAN overfits once the discontinuous bump $U(\mathbf{x})$ is introduced  and its test loss eventually explodes; the deeper MLP is more stable but lags behind GMNM by 1–2 orders of magnitude.  

• Scalability. Even after adding extreme local structure (large $c$ in level 4), GMNM continues to improve, confirming that extra Gaussians can be recruited to model localised detail without harming the global fit.

These observations mirror the 1-D study and suggest that the locality of Gaussians, combined with learnable mixing weights, provides an efficient multi-resolution basis that standard MLP activations struggle to emulate.


\subsection{Solving a Poisson PDE}\label{sec:pde}

Following the experimental protocol in KAN~\cite{liu2024kan}, we test GMNM on a two–dimensional Poisson equation (Eq.~\eqref{eq:pde}).  
The task is a standard benchmark for function approximation under partial supervision: the ground–truth function $f(x,y)$ is never observed; the model only sees its Laplacian $\nabla^{2}f(x,y)$ and the boundary values.

The training loss has two terms:  
(i) the mean–squared error (MSE) between the predicted and true boundary values;  
(ii) the MSE between the predicted Laplacian $\nabla^{2}\text{NN}(x,y)$ and the analytical Laplacian $\nabla^{2}f(x,y)$.  
We compare GMNM with Physics-Informed Neural Networks (PINNs)~(\cite{raissi2017physics}) and KAN.

\begin{equation}\label{eq:pde}
\begin{gathered}
\nabla^{2}f(x,y)=-2\pi^{2}\sin(\pi x)\sin(\pi y),\\
f(-1,y)=f(1,y)=f(x,-1)=f(x,1)=0,\\
f(x,y)=\sin(\pi x)\sin(\pi y).\\
\end{gathered}
\end{equation}

\begin{figure}[h]
  \centering
  \includegraphics[scale=0.35]{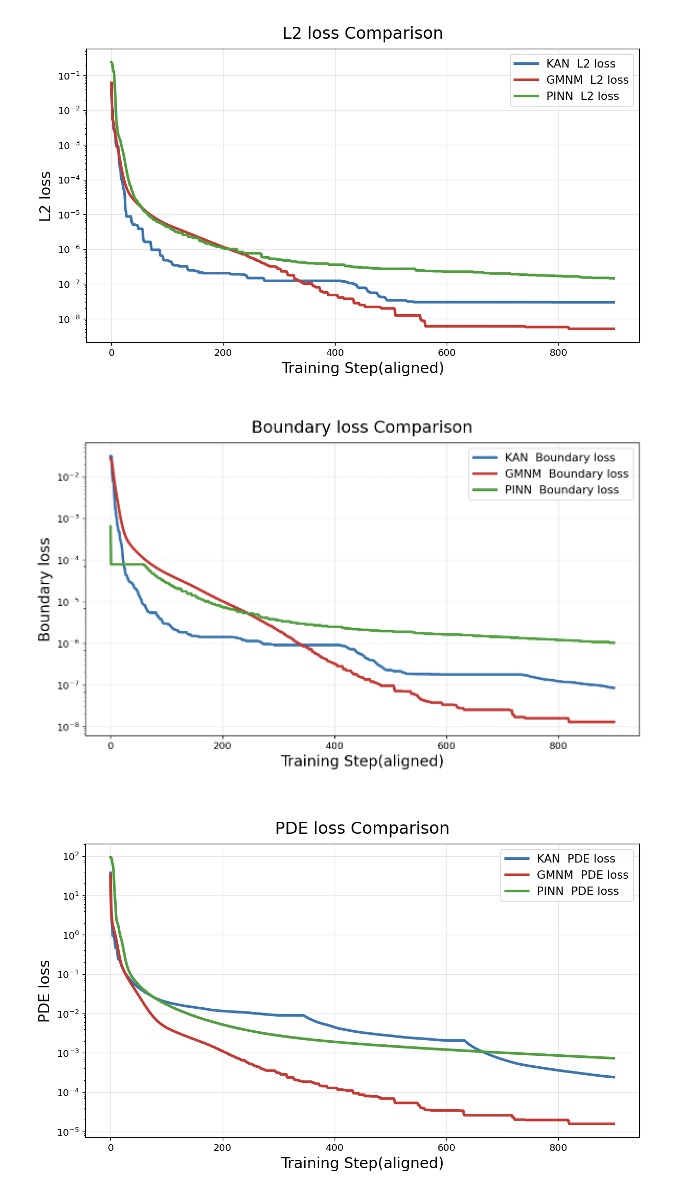}
  \caption{Poisson equation results.  
  Top: total L2 error over training steps.  
  Middle: boundary MSE.  
  Bottom: PDE residual MSE.  
  Lower curves indicate better performance.}
  \label{fig:pde}
\end{figure}

Figure~\ref{fig:pde} plots the learning curves.  
The red line is GMNM, the blue line is KAN, and the green line is the MLP-based PINN.  
We report: (a) the total L2 error of the predicted function, (b) the boundary loss, and (c) the PDE residual.  
Table~\ref{tab:pde} lists the number of trainable parameters and the minimum L2 error achieved by each model.  
GMNM attains the lowest error while using fewer parameters than the baseline networks.

\begin{table}[h]
  \centering
  \caption{Model size and best L2 error on the Poisson problem.}
  \label{tab:pde}
  \begin{tabular}{lccc}
    \toprule
    Model & Architecture & \# Params & Min.~L2 \\
    \midrule
    KAN   & [2, 10, 10, 1]        & 10\,600 & $3.6\times10^{-8}$ \\
    PINN  & [2, 50, 50, 50, 1]   & 5\,301  & $1.5\times10^{-7}$ \\
    \textbf{GMNM}  & \textbf{AGP size 900} & \textbf{4\,500}  & $\mathbf{5.1\times10^{-9}}$ \\
    \bottomrule
  \end{tabular}
\end{table}

The Poisson experiment demonstrates that GMNM can incorporate differential constraints efficiently and achieves an order of magnitude lower error than KAN and PINN.  
While GMNM handles the Mahalanobis distance well in low-dimensional settings, extending this advantage to high-dimensional tasks such as image classification or time-series forecasting remains an open problem.  
The next subsection investigates these scenarios in detail.

\section{Experiments: GMNM in Neural Network}\label{sec:GMNM-cnn}

In the previous section, we demonstrated the superior approximation capabilities of GMNM. However, in many practical applications—such as classification—pattern recognition is equally essential for a model’s performance. In this section, we integrate GMNM into various neural network architectures to showcase its generalizability and practical utility. Through a series of experiments, we aim to provide a comprehensive and systematic analysis of the impact GMNM has on neural networks across diverse tasks.
\textbf{To ensure that any observed performance improvements can be attributed directly to GMNM, we adopt minimalistic network designs and standardized training settings.} 

The experiments are conducted using the simplest possible neural network configurations and include the following tasks: (1) MNIST \cite{deng2012mnist} image classification, (2) CIFAR-10 and CIFAR-100 \cite{krizhevsky2009learning} image classification, (3) CIFAR-10 and CIFAR-100 image generation, and (4) time series forecasting. Our primary focus is on image-based tasks, i.e., experiments (1) to (3), which involve networks composed of convolutional layers, attention mechanisms, and standard fully connected layers. In experiment (4), we assess GMNM’s compatibility with LSTM layers by designing a time series forecasting task.

Across all experiments, we ensure that the models have a similar and relatively small number of trainable parameters. The architectures are kept as simple as possible, and the training objectives and procedures are consistent throughout. This approach allows us to isolate the effect of GMNM by directly comparing model performance with and without the GMNM module. Such a design helps confirm that the improvements are not due to differences in model size, training methodology, or architecture. Following the methodology of \cite{lu2025efficient, lu2025effective}, the GMNM parameters $\mu$ are fixed after initialization.

\subsection{Image Classification}\label{IC}

\begin{figure*}[t]
  \centering
\includegraphics[width=0.9\textwidth]{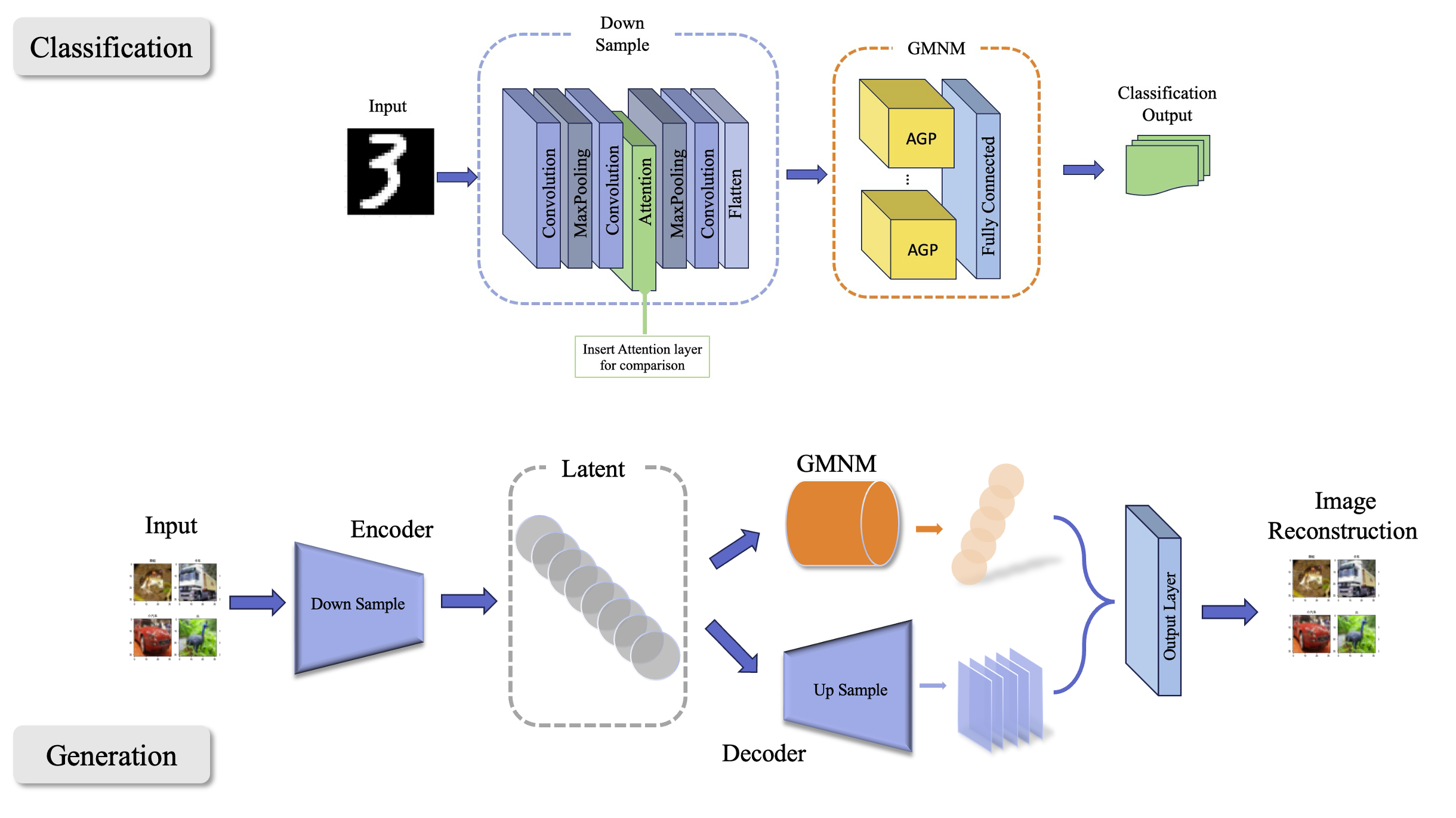}
  \caption{Architecture for Classification Tasks and Generation Tasks. Top (Classification):   
  The green block shows the optional attention layer, while the yellow blocks denote GMNM modules that model channel-wise interactions.
  Bottom (Generation): Architecture follow auto-encoder frame work with convolution down sample and convolution transpose up sample.  Image down sample into a latent vector via the encoder. The latent features also separately feed to the GMNM module. The outputs GMNM module element-wise multiplication operation with decoder output. These fused features are passed through a output  layer(convolution) for the final images reconstructed. }
  \label{fig:cnn-arch}
\end{figure*}

The experimental configurations for image classification are illustrated in top part of Fig.\ref{fig:cnn-arch}. In these experiments, GMNM is employed as the final classification head. This setup requires GMNM not only to approximate the output distribution but also to backpropagate gradients effectively, enabling the preceding network components to perform pattern recognition based on the provided image label information. We compare the performance of three model variants:1.A baseline convolutional neural network (CNN),2.CNN enhanced with an attention block (CNN+Att), and 3.CNN incorporating GMNM as the classification head (CNN+GMNM).
\begin{table}[h]
  \centering
  \caption{Parameter count and training speed of classification experiment. Model for both Cifar 10 and 100 are unchanged. Some convolution layers in GMNM integrated models are tuned smaller to keep similar parameter count.}
  \label{tab:cnn}
  \begin{tabular}{lccc}
    \toprule
    Model & Params  &Params & Speed \\
          & MNIST & Cifar &  Cifar\\
    \midrule
    CNN                    & 28\,586 & 209\,098 & 3ms/step \\
    CNN+Att                & 32\,874 & 213\,386 & 4ms/step  \\
    \textbf{CNN+GMNM}               & \textbf{29\,930} & \textbf{200\,618} & \textbf{3ms/step} \\
    \bottomrule
  \end{tabular}
\end{table}

Table~\ref{tab:clas_comp} presents the minimum training and test losses averaged over six runs (each with 40 training epochs), while Table~\ref{tab:cnn} details the parameter counts and per-step training speed for each model variant.

\textbf{Key observations:}
\begin{itemize}
    \item Models incorporating attention layers achieve lower training loss but exhibit the worst test loss, indicating a tendency to overfit in classification tasks.  
    \item CNN+GMNM consistently yields the lowest test loss and also improves training loss with similar or even fewer parameters compared to the other models.  
    \item GMNM does not introduce any noticeable slowdown in training speed
    \item Across all datasets—MNIST, CIFAR-10, and CIFAR-100—GMNM consistently enhances the performance of the neural networks.
\end{itemize}

These results suggest that GMNM reduces overfitting while boosting accuracy, making it a practical drop-in component for CNNs.

\begin{table}[htbp]
\small
\centering
\caption{Minimum loss of different methods on MNIST, CIFAR-10, and CIFAR-100 for classification task.}
\label{tab:clas_comp}
\begin{tabular}{lllc}
\toprule
Dataset    & Method                   & Loss Type & Min Loss   \\
\midrule
\multirow{4}{*}{MNIST} 
  & CNN                      & Train & 2.812e-03 \\
  & CNN + GMNM             & Train & 1.837e-03 \\
  & CNN + Attention          & Train & 4.979e-04 \\
  & CNN + GMNM + Attention   & Train & 6.220e-04 \\
\cmidrule{2-4}
  & CNN                      & Test  & 8.063e-03 \\
  & \textbf{CNN + GMNM}    & \textbf{Test}  & \textbf{6.081e-03} \\
  & CNN + Attention          & Test  & 9.696e-03 \\
  & CNN + GMNM + Attention   & Test  & 8.122e-03 \\
\midrule
\multirow{4}{*}{CIFAR-10}
  & CNN                      & Train & 0.088     \\
  & CNN + GMNM               & Train & 0.045     \\
  & CNN + Attention          & Train & 0.036     \\
  & CNN + GMNM + Attention   & Train & 0.022     \\
\cmidrule{2-4}
  & CNN                      & Test  & 0.143     \\
  & \textbf{CNN + GMNM}  & \textbf{Test}  & \textbf{0.138} \\
  & CNN + Attention          & Test  & 0.170     \\
  & CNN + GMNM + Attention   & Test  & 0.165     \\
\midrule
\multirow{4}{*}{CIFAR-100}
  & CNN                      & Train & 0.028     \\
  & CNN + GMNM               & Train & 0.025     \\
  & CNN + Attention          & Train & 0.020     \\
  & CNN + GMNM + Attention   & Train & 0.020     \\
\cmidrule{2-4}
  & CNN                      & Test  & 0.035     \\
  & \textbf{CNN + GMNM}   & \textbf{Test}  & \textbf{0.033} \\
  & CNN + Attention          & Test  & 0.038     \\
  & CNN + GMNM + Attention   & Test  & 0.037     \\
\bottomrule
\end{tabular}
\end{table}

\subsection{Image Generation}\label{IG}

The experimental configurations for image generation are depicted in bottom part of Fig.~\ref{fig:cnn-arch}. In these experiments, we adopt a basic autoencoder as the backbone architecture. After encoding, the bottleneck layer produces a latent feature vector, which serves as the input to GMNM. GMNM then outputs a one-dimensional vector that is applied channel-wise as a multiplicative factor to the feature maps immediately before the final image reconstruction layer. Intuitively, we hypothesize that images can be viewed as combinations of basis images, and GMNM is employed to reweight these combinations accordingly.  Similar to the classification experiments, we compare the following models: 1.A simple autoencoder (AE) as the baseline; 2.An autoencoder with an attention block (AE+Att); 3.An autoencoder with GMNM (AE+GMNM); and 4.An autoencoder with both attention and GMNM (AE+Att+GMNM). 

\begin{table}[h]
  \centering
  \caption{Parameter count and training speed of generation tasks for Cifar10 and Cifar100. Model for both Cifar 10 and 100 are unchanged.}
  \label{tab:gen}
  \begin{tabular}{lcc}
    \toprule
    Model   &Params & Speed \\
    \midrule
    AE                     & 2\,813\,167 & 23ms/step \\
    AE+Att              & 3\,224\,687  & 25ms/step \\
    \textbf{AE+GMNM}                & \textbf{2\,778\,723}  & \textbf{17ms/step}  \\
    AE+ATT+GMNM           & 3\,190\,243  & 20ms/step \\
    \bottomrule
  \end{tabular}
\end{table}

\begin{figure}[htbp]
    \centering
    \includegraphics[scale=0.24]{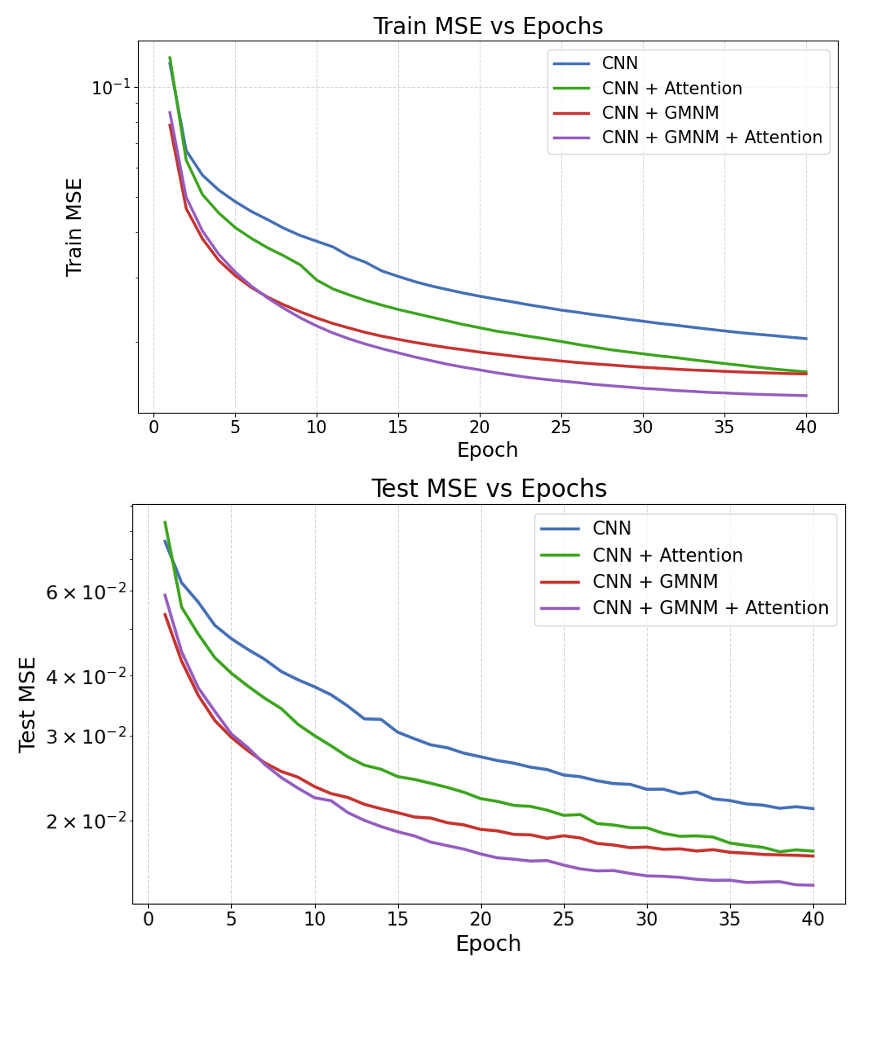}
    \caption{L2 loss for Cifar100 image generation experiment. Top: training loss; Bottom: test loss. }
    \label{fig:c100-gen}
\end{figure}

\begin{table}[h]
  \centering
  \caption{Minimum L2 loss of Cifar 10 and 100 image generation task.}
  \label{tab:gen2}
  \begin{tabular}{lcc}
    \toprule
    Model   &Cifar 10 & Cifar 100 \\
       &Train/Test & Train/Test \\
    \midrule
    AE                     & (0.0187)/(0.0190) & (0.020)/(0.020)  \\
    AE+Att              & (0.0160)/(0.0162)  & (0.0165)/(0.0170) \\
    AE+GMNM                & (0.0152)/(0.0153)  & (0.0161)/(0.0167)  \\
    \textbf{AE+ATT+GMNM}           & \textbf{(0.0148)/(.0151)} & \textbf{(0.0141)/(0.0144)}  \\
    \bottomrule
  \end{tabular}
\end{table}
Figure~\ref{fig:c100-gen} presents the average training and test losses across six runs (each with 40 training epochs) on dataset cifar 100. Cifar10 (see supplementary materials) and Cifar100 share similar lost descend. The minimum MSE lost is given by Table ~\ref{tab:gen2}. Table~\ref{tab:gen} summarizes the parameter counts and per-step training speed for each model configuration. The key observations in the image generation experiments closely align with those from the classification tasks in the previous section:

\begin{itemize}
    \item GMNM+ Models consistently achieve the lowest test loss and also show improved training loss, while maintaining the smallest parameter count among all variants.
    \item The integration of GMNM has no noticeable impact on training speed.
    \item Across both CIFAR-10 and CIFAR-100 datasets, GMNM consistently enhances neural network performance.
    \item GMNM + attention mechanisms further improves model performance, suggesting their complementary benefits.
\end{itemize}

\subsection{Time–series Forecasting}\label{sec:GMNM-lstm}
In this finaly experiment, we embed GMNM into a lightweight LSTM to test whether it improves sequential modelling.

Four latent signals are generated as
\begin{equation}\label{eq:ts}
\begin{aligned}
x_i(t) &= a_i\sin t + b_i , \qquad i=1,\dots,4,\\
y(t) &= x_3(t)\,x_4(t-0.1)
      -x_3(t-0.5)\,x_1(t-0.1)\\
     &\phantom{{}=}
      +x_4(t)\,x_3(t)
      -x_2(t-0.5)\,x_1(t-0.2),
\end{aligned}
\end{equation}
with $t\in[0.5,100]$.  
We draw 10,000 samples, use 80 \% for training and 20 \% for test, and feed the network the last ten time steps of the four $x_i$ as input.

We compare  
(i) LSTM with 3 hidden units,  
(ii) the same LSTM plus an GMNM layer of 100 components, and  
(iii) a larger LSTM with 32 units that has similar parameter count to the GMNM variant (more details can be found in supplementary materials).  
All networks are trained for 50 epochs and each experiment is repeated 20 times.


\begin{table}[h]
  \centering
  \caption{L2 lost comparison of LSTM variants for time series forcaseting. LSTM+GMNM use less LSTM unit but achieve one magnitude lower L2 lost.}
  \label{tab:lstm-loss}
  \begin{tabular}{lcc}
    \toprule
    Model   &Train Lost & Test Lost\\
    \midrule
    LSTM(3)                     &$7\times 10^{-2}$& $6\times10^{-2}$\\
    \textbf{LSTM(3)+GMNM }             & $\mathbf{4\times10^{-5}}$  & $\mathbf{2\times10^{-5}}$\\
    LSTM(32)               & $3\times10^{-4}$  & $7\times10^{-4}$  \\
    \bottomrule
  \end{tabular}
\end{table}


Table~\ref{tab:lstm-loss} shows that replacing the larger LSTM with a small LSTM plus GMNM lowers both training and test loss by an order of magnitude, indicating that GMNM captures non-linear interactions more effectively than additional recurrent units.

\section{Conclusion, Limitation and Discussions}

In this work, we have introduced the Gaussian Mixture-Inspired Nonlinear Modules (GMNM), a novel neural architecture that effectively integrates Gaussian Mixture Models directly into neural networks without requiring specialized algorithms such as the Expectation-Maximization. Experimental results demonstrate that GMNM exhibits superior function approximation capability and significantly enhances existing neural architectures, highlighting its promising potential for diverse applications. This intrinsic nonlinearity module enriches modeling expressiveness, consistently improve neural network performance across all experiments.

Nevertheless, GMNM faces practical limitations. Increasing approximation accuracy typically necessitates expanding the number of Gaussian components, leading to wider rather than deeper architectures. Addressing these scalability concerns will require further engineering improvements. Theoretical and experimental works are required to systematic address how GMNM should be used in neural network.

Finally, we propose several conjectures based on our observations. Each Gaussian component in GMNM responds to specific input patterns, much like how human brain are locally specialized for particular cognitive functions. In several experiments, we also employed non-trainable $\mu$ values, which still led to effective model performance. These findings point to a compelling hypothesis: in large-scale models—such as models with billions of parameters—if the parameters are appropriately initialized and sufficiently cover the parameter space (i.e., exhibit ergodicity), a significant portion of the model may already possess useful representational capacity, potentially reducing the need for extensive training. This insight could inform future strategies for improving training efficiency in large neural systems.

\bibliographystyle{unsrtnat}
\bibliography{references}  






\end{document}